\documentclass[conference]{IEEEtran}
\usepackage{amsmath, amssymb, amsfonts}
\usepackage{fontenc}
\usepackage{graphicx}
\usepackage{color}
\usepackage[rgb]{xcolor}
\usepackage{longtable}
\usepackage[colorlinks=true]{hyperref}
\usepackage{float}
\usepackage{epsfig}
\usepackage{epstopdf}
\usepackage{algorithm}
\usepackage{algorithmic}
\usepackage{rotating}
\usepackage{tikz}
\usetikzlibrary{calc,through,backgrounds}
\usetikzlibrary{arrows,automata,positioning,fit}
\usetikzlibrary{decorations.pathreplacing} 

\newcommand{\vect} [1] {\boldsymbol{#1}}
\hypersetup{
    bookmarks=true,         % show bookmarks bar?
    colorlinks=true,       % false: boxed links; true: colored links
    linkcolor=red,          % color of internal links
    citecolor=blue,        % color of links to bibliography
    filecolor=blue,      % color of file links
    urlcolor=red           % color of external links
}

\ifCLASSINFOpdf
\else
\fi
\begin{document}
%\title{A Bayesian Approach for Combining Classifier and Cluster Ensembles with Privacy Preservation}
\title{A Privacy-Aware Bayesian Approach for Combining Classifier and Cluster Ensembles}


\author{\IEEEauthorblockN{\centering Ayan Acharya\textsuperscript{1}, Eduardo R. Hruschka\textsuperscript{1,2}, and Joydeep Ghosh\textsuperscript{1}}\\
\IEEEauthorblockA{\textsuperscript{1}University of Texas (UT) at Austin, USA \\ \textsuperscript{2}University of Sao Paulo (USP) at Sao Carlos, Brazil}}

% \and
% \IEEEauthorblockN{Homer Simpson}
% \IEEEauthorblockA{Twentieth Century Fox\\
% Email: homer@thesimpsons.com}
% \and
% \IEEEauthorblockN{James Kirk\\ and Montgomery Scott}
% \IEEEauthorblockA{Starfleet Academy\\
% San Francisco, California 96678-2391\\
% Telephone: (800) 555--1212\\


%Telephone: (800) 555--1212, Fax: (888) 555--1212}
\maketitle

\begin{abstract}
%\boldmath
This paper introduces a privacy-aware Bayesian approach that combines ensembles of classifiers and clusterers to perform semi-supervised and transductive learning.
We consider scenarios where instances and their classification/clustering results are distributed
across different data sites and have sharing restrictions. As a special case,  
the privacy aware computation of the model when instances of the target data are distributed across different data sites, is also discussed.
Experimental results show that the proposed approach can provide good classification accuracies while adhering to the data/model sharing constraints.

\end{abstract}


% \fi
\IEEEpeerreviewmaketitle

\section{Introduction}

Extracting useful knowledge from large, distributed data repositories can be a very difficult task 
when such data cannot be directly centralized or unified as a single file or database
due to a variety of constraints.
% As in much of parallel processing, early work on distributed data mining largely
% focused more on technical constraints such as low communication bandwidths or limited central storage.
%More 
Recently, there has been an emphasis 
on how to obtain high quality information from distributed sources via statistical modeling
while simultaneously adhering to restrictions on the {\em nature} of the data
or models to be shared, due to data ownership or privacy issues.
Much of this work has appeared under the moniker of ``privacy-preserving data mining''.

Three of the most popular approaches to privacy-preserving data mining techniques are:
(i) query restriction to solve the inference problem in databases \cite{faja02}
(ii) subjecting individual records or attributes to a ``privacy preserving" randomization operation 
and subsequent recovery of the original data \cite{agsr00},
(iii) using cryptographic techniques for secure two-party or multi-party communications \cite{pink02}.
% The first method is difficult and manually intensive, while the latter two
% approaches are largely restricted to vector data and 
% %and also miss the key data mining goal of characterizing/modeling data 
% %and finding useful patterns in it, rather than simply restoring  data.
% are applicable only in settings where a central party is
% collecting individual records that need to be protected.
Meanwhile, the notion of privacy has expanded substantially over the years. 
% Early notions were often black-and-white, 
% for example, under HIPAA (Health Insurance Portability and Accountability Act) rules, certain fields can be readily shared 
% while others cannot be revealed under any circumstance. 
Approaches such as $k$-anonymity and $l$-diversity \cite{makg06} focused 
on privacy in terms of indistinguishableness of one record from others under allowable queries. More recent approaches such as 
differential privacy \cite{dwle09} tie the notion of privacy to its impact on a statistical model. 
% In this paper, privacy is dealt with in a statistical 
% manner based on a quantifiable information-theoretic formulation \cite{megh03}. 
% Informally speaking, a dataset (or a subset of the same) 
% is represented using a parametric model, and the lower the probability of recovering the actual data records given such a model, the more 
% privacy-aware that model is.

The larger body of distributed data mining techniques developed so far have focused on simple classification/clustering algorithms
or on mining association rules \cite{agag01,chsw96,evsr02,lipi00}. 
Allowable data partitioning is also limited, typically to {\em vertically partitioned}
%(different sites contain different attributes/features of a common set of records/instances) 
or {\em horizontally partitioned} 
% (\textit{i.e.}, the instances are distributed amongst
% the sites, which record the same set of features for
% each instance) 
data \cite{dhmo99}. 
These techniques typically do not specifically address privacy issues, other than through encryption \cite{vacl03}. This is also true of earlier, data-parallel methods \cite{dhmo99} that are susceptible to privacy breaches, and also need a central planner that dictates what algorithm runs on each site. 
%(IF SPACE PERMITS, WE COULD ADD SOME MORE RECENT REFERENCES FOR PRIVACY-PRESERVING CLUSTERING, CLASSIFICATION AND ASSOCIATION)
In this paper, we introduce a privacy-aware Bayesian approach that combines ensembles of classifiers and clusterers and is effective for 
both semi-supervised and transductive learning.
As far as we know, this topic has not been addressed in the literature.
% ; we could not even locate one work that is reasonably close in its application 
% setting and incorporates privacy issues.

The combination of multiple classifiers to generate an ensemble has been proven to be
more useful compared to the use of individual classifiers \cite{oztu08}.
%-- both theoretically and empirically \cite{oztu08}.  
Analogously, several research efforts have shown that cluster ensembles can improve the quality of results as compared to a single clusterer --- {\it e.g.}, see \cite{wasb11} and references therein. 
%Indeed, the potential motivations and benefits for using cluster ensembles are much broader than those for using classifier ensembles, for which improving the predictive accuracy is usually the primary goal. More specifically, cluster ensembles can be used to generate more robust and stable results (compared to a single clustering approach), perform distributed computing under privacy or sharing constraints, or reuse existing knowledge \cite{stgh02b}. 
%Analogously, several research efforts have shown that cluster ensembles can improve the quality of results as compared to a single clustering 
%solution -- {\it e.g.}, see \cite{wasb11} and references therein. Improving the predictive accuracy is often the primary objective 
%in the use of classifier ensemble. However, cluster ensemble
%can not only be used to generate more robust and stable results (compared to a single clustering approach) but also 
%be utilized to perform 
%distributed computing under privacy or sharing constraints, or reuse existing knowledge \cite{stgh02b}. 
Most of the motivations for combining ensembles of classifiers and clusterers are similar to those that hold for the standalone use of 
either classifier or cluster ensembles. However, some additional nice properties can emerge from such a combination. For instance, unsupervised 
models can provide supplementary constraints for classifying new data and thereby improve the generalization
capability of the resulting classifier. 
% From this viewpoint, the underlying assumption is 
% that similar new instances of the target set are more likely to share the same class label. 
% Also, they can be useful for designing learning methods 
% that are aware of the possible differences between training and target instances, thus being particularly interesting for applications in which concept 
% drift may take place. 
Having this motivation in mind, a Bayesian approach to combine cluster and classifier ensembles in a privacy-aware setting is presented. 
We consider that a collection of instances and their clustering/classification algorithms reside in different data sites. 
% This scenario is common in many practical problems. 
% For example, the data sites can represent parties that are a group of 
% banks, with their own sets of customers, who would like to have a better insight into the  behavior of the entire customer population without 
% compromising the privacy of their individual customers. 

The idea of combining classification and clustering models has been introduced in the 
algorithms described in 
%Bipartite Graph-based Consensus Maximization (\textbf{BGCM}) algorithm 
\cite{galf09,achg11}. 
%and in the \textbf{C\textsuperscript{3}E}  algorithm \cite{achg11}.
However, these algorithms do not deal with privacy issues. Our probabilistic framework provides an alternative approach 
to combining class labels with cluster labels under conditions where sharing of individual records across data sites is not permitted. This 
soft probabilistic notion of privacy, based on a quantifiable information-theoretic formulation, has been discussed in detail in \cite{megh03}.

\section{\textbf{BC\textsuperscript{3}E} Framework}
\label{sec:BC3E_Framework}

\subsection{Overview}
\label{Overview}
%The framework that combines classifier and cluster ensembles to generate a more consolidated 
%classification is depicted in Fig. \ref{fig:1}. %\ref{frame}
Consider that a classifier ensemble previously induced from training data is employed to generate a set of class labels for 
every instance in the target data. Also, a cluster ensemble is applied to the target data to provide sets of cluster labels. These class/cluster labels provide the inputs to Bayesian Combination of Classifier and Cluster Ensembles (\textbf{BC\textsuperscript{3}E}) algorithm.

\subsection{Generative Model}
\label{sec:BC3E}

Consider a target set $\mathcal{X}=\{\mathbf{x}_{n}\}_{n=1}^{N}$ formed by $N$ unlabeled instances. 
Suppose that a classifier ensemble composed of $r_{1}$ classification models has produced $r_{1}$ class labels (not necessarily different) for every instance $\mathbf{x}_{n}\in\mathcal{X}$. Similarly, consider that a cluster ensemble comprised of $r_{2}$ clustering algorithms has generated cluster labels for every instance in the target set. 
% Note that (i) the number of clusters is not an input, and it is quite possible that different clustering algorithms partition the 
% target data into different numbers of clusters; and (ii) cluster labels are categorical variables. 
Note that the cluster labeled as \textit{1} in a given data partition may not align with the cluster numbered \textit{1} in another partition, 
and none of these clusters may correspond to class \textit{1}. Given the class and cluster labels, the objective is to come up with 
refined class probability distributions $\{\vect{\theta}_{n}\}_{n=1}^{N}$ of the target set instances. 
%From this point of view, 
%the cluster ensemble provides supplementary constraints for classifying the instances of $\mathcal{X}$, 
%with the rationale that similar instances are more likely to share the same class label.
To that end, assume that there are $k$ classes, which are denoted by $C=\{C_{i}\}_{i=1}^{k}$. 
The observed class and cluster labels are denoted by $\vect{X}=\{\{w_{1nl}\},\{w_{2nm}\}\}$ where $w_{1nl}$ 
is the class label of the $n^{\text{th}}$ instance for the $l^{\text{th}}$ classifier and $w_{2nm}$ is the cluster label 
assigned to the $n^{\text{th}}$ instance by the $m^{\text{th}}$ clusterer. 
A generative model is proposed to explain the observations $\vect{X}$, where each instance $\mathbf{x}_{n}$ 
has an underlying mixed-membership to the $k$ different classes. Let $\boldsymbol{\theta}_{n}$ denote the latent 
mixed-membership vector for $\mathbf{x}_{n}$. 
It is assumed that $\boldsymbol{\theta}_{n}$ -- a discrete probability distribution over the $k$ classes -- is sampled from a 
Dirichlet distribution, with parameter $\boldsymbol{\alpha}$. Also, for the $k$ classes (indexed by $i$) 
and $r_{2}$ different base clusterings (indexed by $m$), we assume a multinomial distribution $\boldsymbol{\beta}_{mi}$ over the cluster labels.
If the $m^{\text{th}}$ base clustering has $k^{(m)}$ clusters, $\boldsymbol{\beta}_{mi}$ is of dimension $k^{(m)}$ and
$\sum_{j=1}^{k^{(m)}}\boldsymbol{\beta}_{mij}=1$.  
The generative model can be summarized as follows. For each $\mathbf{x}_n\in\mathcal{X}$:
\begin{enumerate}
 \item Choose $\vect{\theta}_{n}\sim\text{Dir}(\vect{\alpha})$.
%where $\vect{\alpha}$ is the parameter of a Dirichlet distribution of dimension $k$.
 \item $\forall l\in\{1,2,\cdots,r_{1}\}$, choose $w_{1nl}\sim \text{multinomial}(\vect{\theta}_{n})$. 
 \item $\forall m\in\{1,2,\cdots,r_{2}\}$. 
 \begin{enumerate}
 \item Choose $\vect{z}_{nm}\sim \text{multinomial}(\boldsymbol{\theta}_{n})$ 
where $\vect{z}_{nm}$ is a vector of dimension $k$ with only one component being unity and others being zero.
 \item Choose $w_{2nm}\sim \text{multinomial}(\boldsymbol{\beta}_{r\vect{z}_{nm}})$.
 \end{enumerate}
\end{enumerate}

If the $n^{\text{th}}$ instance is sampled from the $i^{\text{th}}$ class in the $m^{\text{th}}$ base clustering (implying $z_{nmi}=1$), 
then its cluster label will be sampled from the multinomial distribution
$\boldsymbol{\beta}_{mi}$. 
% This representation is natural because, given the true class, certain cluster labels are more likely than others. 
% It is also general as it allows multimodal classes and does not force a one-to-one correspondence between classes and clusters.  
Modeling of the classification results from $r_{1}$ different classifiers for the $n^{\text{th}}$ instance is straightforward: 
the observed class labels ($\{w_{1nl}\}$) are assumed to be sampled from the latent mixed-membership vector $\boldsymbol{\theta}_{n}$.
%The corresponding graphical model is shown in Fig. \ref{BC3E_fig}. 
In essence, the posteriors of $\{\boldsymbol{\theta}_{n}\}$ are expected to get more accurate in an effort to 
explain both classification and clustering results (\emph{i.e.} $\vect{X}$) in the same framework.
\textbf{BC\textsuperscript{3}E} derives its inspiration from the mixed-membership na\"{\i}ve Bayes model \cite{shba11}. 
% Theoretical motivations for working with
% such models can be found in \cite{wasb11,shba11} and are omitted here due to space constraints.
%details of which are omitted here due to space constraints. THIS SENTENCE WAS RATHER STRANGE...

% \begin{figure}
% \begin{center}
% \input{BC3E_fig.tex} 
% \caption{Graphical Model for BC\textsuperscript{3}E}
% \label{BC3E_fig}
% \end{center}
% \end{figure}

To address the log-likelihood function of \textbf{BC\textsuperscript{3}E}, let us denote the set of hidden variables by 
$\vect{Z}= \{\{\vect{z}_{nm}\}, \{\vect{\theta}_{n}\}\}$. The model parameters can conveniently be represented 
by $\vect{\zeta}_{0}=\{\vect{\alpha},\{\vect{\beta}_{mi}\}\}$.
Therefore, the joint distribution of the hidden and observed variables can be written as:
\begin{eqnarray}
\label{incmpllkhd}
 p(\mathbf{X},\vect{Z}|\vect{\zeta}_{0})&=&
\prod_{n=1}^{N}p(\boldsymbol{\theta}_{n}|
\boldsymbol{\alpha})
\prod_{l=1}^{r_{1}}p(w_{1nl}|\vect{\theta}_{n})\nonumber\\
&&\prod_{m=1}^{r_{2}}p(\vect{z}_{nm}|\boldsymbol{\theta}_{n})
p(w_{2nm}|\boldsymbol{\beta}, \vect{z}_{nm})
\end{eqnarray}

In theory, inference and estimation with the proposed model could be performed 
by maximizing the log-likelihood in Eq. (\ref{incmpllkhd}) -- using 
the \emph{Expectation Maximization} family of algorithms \cite{delr77}. However, the coupling between
$\boldsymbol{\theta}$ and $\boldsymbol{\beta}$ makes
the exact computation in the summation over the classes intractable in general \cite{blnj03}. Therefore, 
inference and estimation is performed using Variational Expectation Maximization (\textbf{VEM}) \cite{jogj99}.
% \footnote{One can also adopt Laplace approximation or 
% MCMC methods~\cite{jordan99}.} 

\subsection{Approximate Inference and Estimation}
\label{sec:infnest}

% \textbf{VEM} does not only make the inference and estimation computationally feasible but also facilitates a distributed computation 
% where privacy about the class/cluster labels can be preserved
%, as will be discussed later in Section \ref{sec:privpres}. 
\subsubsection{Inference}
\label{inference}
To obtain a tractable lower bound on the observed log-likelihood, we specify a fully factorized distribution to approximate the true posterior of the hidden variables:
\begin{equation}
\label{factorized}
 q(\vect{Z}|\{\vect{\zeta}_{n}\}_{n=1}^{N})
=\prod_{n=1}^{N}q(\vect{\theta}_{n}|\vect{\gamma}_{n})
\prod_{m=1}^{r_{2}}q(\vect{z}_{nm}|\vect{\phi}_{nm})
\end{equation}
where  $\vect{\theta}_{n}\sim\text{Dir}(\vect{\gamma}_{n})\text{ }\forall n\in\{1,2,\cdots, N\}$, $\vect{z}_{nm}\sim \text{multinomial}(\vect{\phi}_{nm}) 
\text{ }\forall n\in\{1,2,\cdots, N\}$ and $\forall m\in\{1,2,\cdots,r_{2}\}$, and 
$\vect{\zeta}_{n}=\{\vect{\gamma}_{n},\{\vect{\phi}_{nm}\}\}$, which is the 
set of variational parameters corresponding to the $n^{\text{th}}$ instance.
Further, $\boldsymbol{\alpha}=(\alpha_{i})_{i=1}^{k}$, 
$\boldsymbol{\gamma}_{n}=(\gamma_{ni})_{i=1}^{k}\text{ }\forall n$,
and $\boldsymbol{\phi}_{nm}=(\phi_{nmi})_{i=1}^{k}\text{ }\forall n,m$; where the components of 
the corresponding vectors are made explicit. Using Jensen's inequality, a lower bound on the 
observed log-likelihood can be derived:
\begin{eqnarray}
\label{lwbound}
 \text{log} [p(\vect{X}|\vect{\zeta}_{0})] &\ge&
\mathbf{E}_{q(\vect{Z})}\left[\text{log} [p(\vect{X},\vect{Z}|\vect{\zeta}_{0})]\right]
+H(q(\vect{Z}))\nonumber\\
&=&\mathcal{L}(q(\vect{Z}))
\end{eqnarray}
where $H(q(\vect{Z}))=-\mathbf{E}_{q(\vect{Z})}[\text{log} [q(\vect{Z})]]$ is the entropy of the variational distribution $q(\vect{Z})$,
and $\mathbf{E}_{q(\vect{Z})}[.]$ is the expectation w.r.t $q(\vect{Z})$.
It turns out that the inequality in (\ref{lwbound}) is due to the non-negative KL divergence
between $q(\vect{Z}|\{\vect{\zeta}_{n}\})$ and $p(\vect{Z}|\vect{X},\vect{\zeta}_{0})$ -- the true posterior of the hidden variables.
Let $\mathbf{\mathcal{Q}}$ be the set 
of all distributions having a fully factorized form as given in (\ref{factorized}). The optimal 
distribution that produces the tightest possible lower bound $\mathcal{L}$ is thus given by:
\begin{eqnarray}
\label{lwboundopt}
 q^{*}&=& \arg\min_{q\in\mathbf{\mathcal{Q}}}\text{KL}(p(\vect{Z}|\vect{X},\vect{\zeta}_{0})||q(\vect{Z})).
% &=&q(\vect{Z}|\{\vect{\zeta}_{n}\}^{*})
% =\arg\max_{q\in\mathbf{\mathcal{Q}}}\mathcal{L}(q(\vect{Z}))\nonumber\\ 
\end{eqnarray}
The optimal value of $\phi_{nmi}$ that satisfies (\ref{lwboundopt}) is given by
%\footnote{See \cite{blnj03} for a detailed calculation.}: 
\begin{equation}
\label{updatephi}
 \phi_{nmi}^{*}\propto \text{exp}(\psi(\gamma_{ni}))\displaystyle\prod_{j=1}^{k^{(m)}}{\beta_{mij}}^{w_{2nmj}} \text{ }\forall n,m,i,
\end{equation}
where, $w_{2nmj}=1$ if the cluster label of the $n^{\text{th}}$ instance in the $m^{\text{th}}$
clustering is $j$ and $w_{2nmj}=0$ otherwise.
Since $\boldsymbol{\phi}_{nm}$ is a multinomial distribution, the updated values of the $k$ components should be normalized to unity. 
Similarly, the optimal value of $\{\gamma_{ni}\}$ that satisfies (\ref{lwboundopt}) is given by:
\begin{eqnarray}
\label{updategamman}
\gamma_{ni}^{*} = \alpha_{i} +\displaystyle\sum_{l=1}^{r_{1}}w_{1nli} +\displaystyle\sum_{m=1}^{r_{2}}\phi_{nmi}
% \mathcal{L}_{[\gamma_{n}]} &=& \displaystyle\sum_{i=1}^{k}\text{log}(\Gamma(\gamma_{ni}))
% -\text{log}(\Gamma(\displaystyle\sum_{i=1}^{k}\gamma_{ni}))\nonumber\\
% &+& \displaystyle\sum_{i=1}^{k}\left[\psi(\gamma_{ni})-\psi(\displaystyle\sum_{i=1}^{k}\gamma_{ni})\right]
% \left[\alpha_{i}
% -\displaystyle\sum_{i=1}^{k}\gamma_{ni}\right.\nonumber\\
% &+&\left.\displaystyle\sum_{l=1}^{r_{1}}\displaystyle\sum_{i=1}^{k}w_{1nli}
% +\displaystyle\sum_{m=1}^{r_{2}}\displaystyle\sum_{i=1}^{k}\phi_{nmi}\right]
\end{eqnarray}
% However, direct update of $\{\vect{\gamma}_{n}\}$ to their optimal values is not possible
% (see \cite{blnj03} for more details) and a numerical method for optimization needs to be 
% used\footnote{We use a Newton-Raphson based update procedure as suggested in \cite{minka03}.}. 
% The part of the lower bound that depends on the variational parameter $\vect{\gamma}_{n}$ is provided in Eq. (\ref{updategamman}). 
% Since $\{\vect{\gamma}_{n}\}$ are the parameters of a Dirichlet distribution and hence always 
% non-negative, this constraint should be maintained while maximizing the objective function in Eq. (\ref{updategamman}).
Note that the optimal values of $\vect{\phi}_{nm}$ depend on $\vect{\gamma_{n}}$ and vice-versa. 
Therefore, iterative optimization is adopted to minimize the lower bound till convergence is achieved.

\subsubsection{Estimation}

For estimation, we maximize the optimized lower bound obtained 
from the variational inference w.r.t the free
model parameters $\vect{\zeta}_{0}$ (by keeping the variational parameters fixed). 
Taking the partial derivative of the lower bound w.r.t $\boldsymbol{\beta}_{mi}$ we have: 
\begin{equation}
\label{updatebeta}
\beta_{mij}^{*}\propto\displaystyle\sum_{n=1}^{N}\phi_{nmi}w_{2nmj}\text{ }\forall j\in{1,2,\cdots,k}
\end{equation}
Again, since $\boldsymbol{\beta}_{mi}$ is a multinomial distribution, the updated values of $k^{(m)}$ components should be normalized to unity.
However, no direct analytic form of update exists for $\vect{\alpha}$ (see \cite{blnj03} for more details) and a numerical 
method for optimization needs to be 
resorted to\footnote{We use a Newton-Raphson based update procedure as suggested in \cite{minka03}.}
% , and a numeric optimization method has to be . 
The part of the objective function that depends on $\vect{\alpha}$ is given by: 
\begin{eqnarray}
\label{updatealpha}
\mathcal{L}_{[\alpha]} &=& N\left[\displaystyle\sum_{i=1}^{k}\text{log}(\Gamma(\alpha_{i}))
-\text{log}(\Gamma(\displaystyle\sum_{i=1}^{k}\alpha_{i}))\right]\nonumber\\
&+& \displaystyle\sum_{n=1}^{N}\displaystyle\sum_{i=1}^{k}\left[\psi(\gamma_{ni})-\psi(\displaystyle\sum_{i=1}^{k}\gamma_{ni})\right]
(\alpha_{i}-1)
\end{eqnarray}
Note that the optimization has to be performed with the constraint $\vect{\alpha}\ge \vect{0}$. Once the optimization in M-step is done, 
E-step starts and the iterative update is continued till convergence.
% The main steps of inference and estimation 
% are concisely presented in Algorithm \ref{algo:LearnBC3E}. 
% 
% \input{algorithm_BC3E.tex}

\section{Privacy Aware Computation}
\label{sec:privpres}

Inference and estimation using \textbf{VEM} allows performing computation without explicitly revealing the class/cluster labels. 
%Consider that the instances of the target set (and, accordingly, their respective cluster/class labels) are distributed across different data sites. 
One can visualize instances, along with their class/cluster labels, arranged in a matrix form so that each data site contains a subset of the matrix entries.
Depending on how the matrix entries are distributed across different sites, three scenarios can arise -- 
i) \textit{Row Distributed Ensemble}, ii) \textit{Column Distributed Ensemble}, and iii) \textit{Arbitrarily Distributed Ensemble}. 

\begin{center}
\begin{figure*}[ht]
\begin{minipage}[b]{0.3\linewidth}
\centering
\includegraphics[bb=0 0 210 204, scale=0.6]{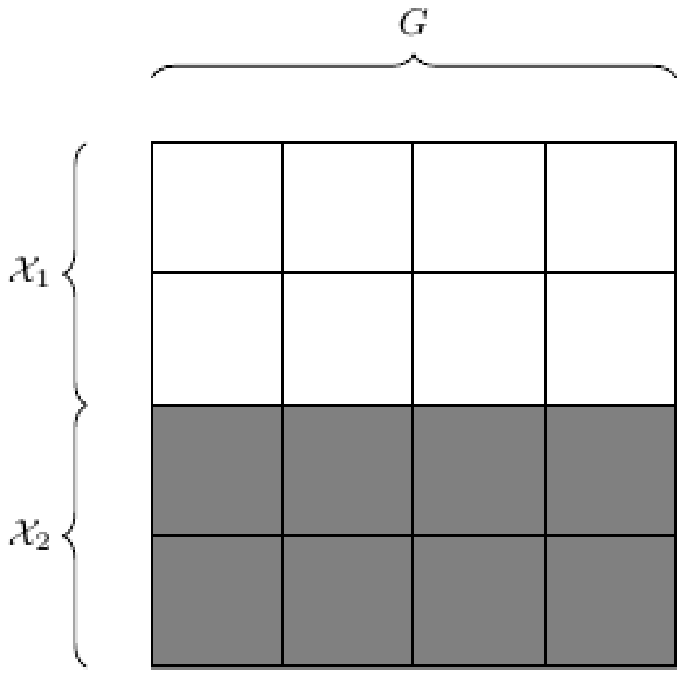}
\caption{Row Distributed Ensemble}
\label{fig:rpart}
\end{minipage}
\hspace{0.02cm}
\begin{minipage}[b]{0.3\linewidth}
\centering
\includegraphics[bb=0 0 205 204,scale=0.6]{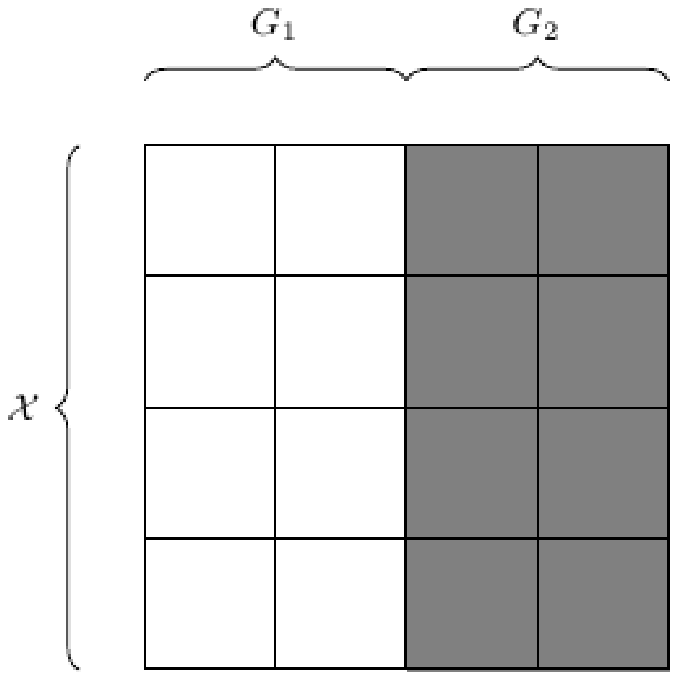}
\caption{Column Distributed Ensemble}
\label{fig:cpart}
\end{minipage}
\hspace{0.02cm}
\begin{minipage}[b]{0.3\linewidth}
\centering
 \includegraphics[bb=0 0 240 316,scale=0.30]{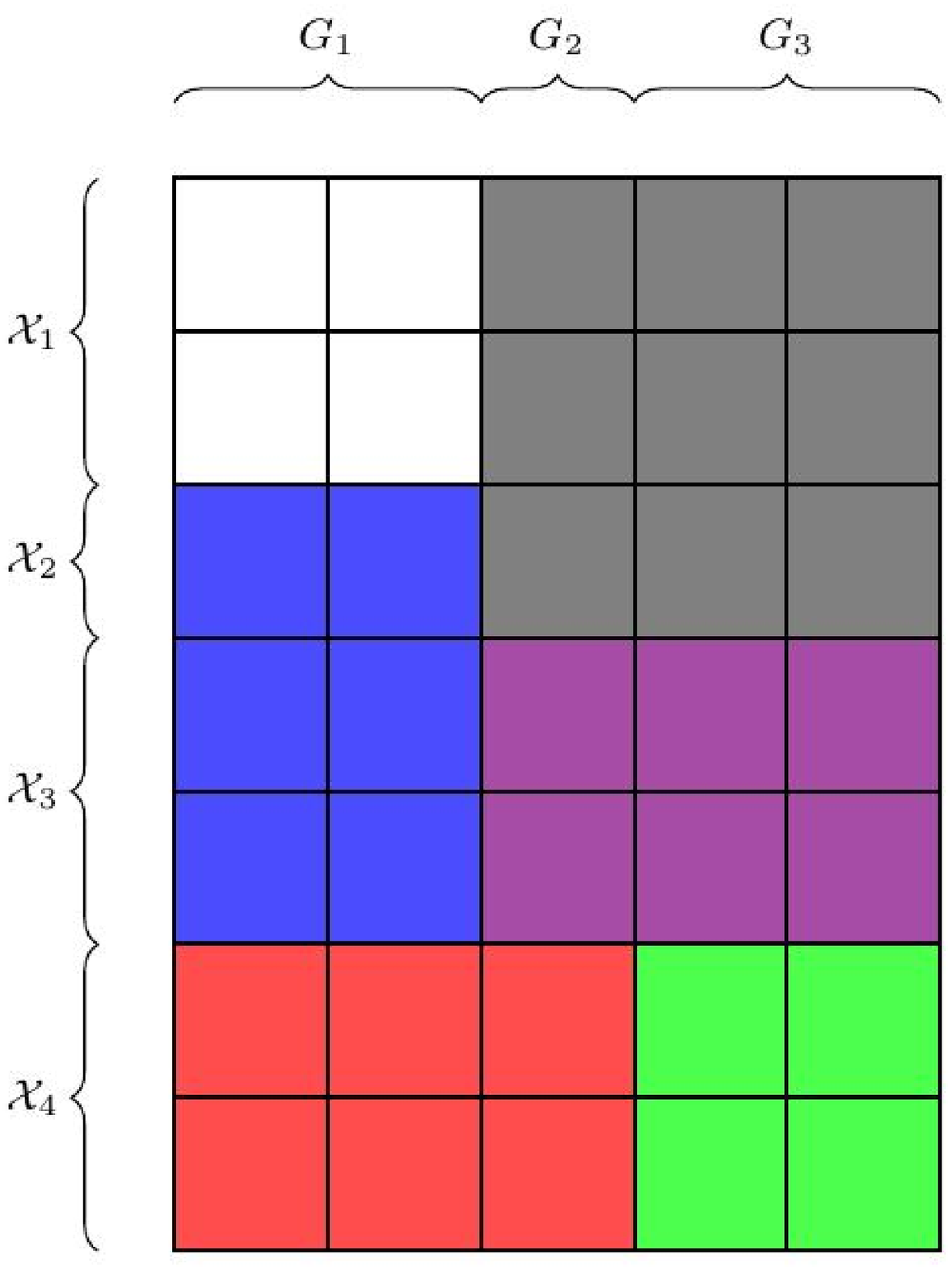}
\caption{Arbitrarily Distributed Ensemble}
\label{fig:arbpart}
\end{minipage}
 \centering
 \includegraphics[bb=0 0 810 177,scale=0.5]{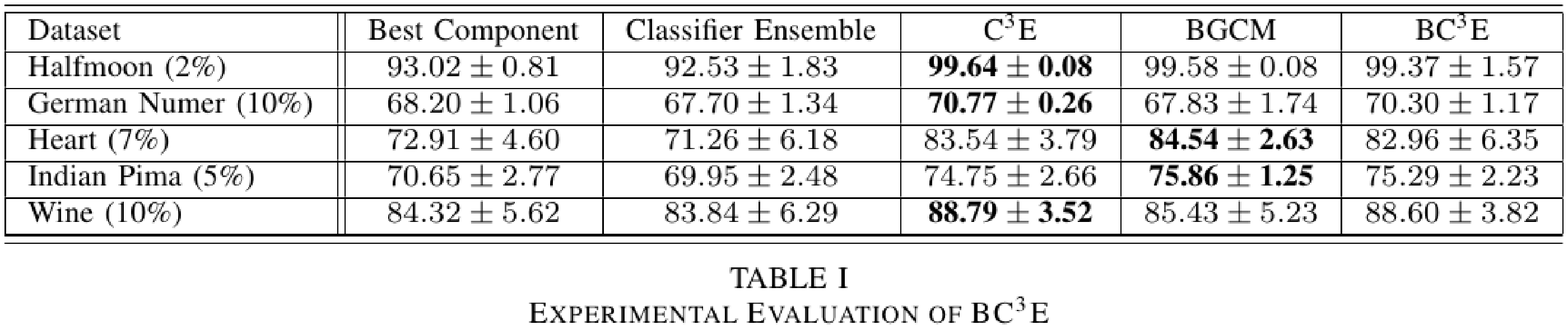}
 % results.png: 1080x236 pixel, 96dpi, 28.57x6.24 cm, bb=0 0 810 177
\end{figure*}
\end{center} 

% \begin{figure*}[!h]
%  \centering
%  \includegraphics[bb=0 0 543 117,scale=0.8]{./results.png}
%  % results.png: 724x156 pixel, 96dpi, 19.15x4.13 cm, bb=0 0 543 117
% \end{figure*}
  
\vspace{-0.9cm}
\subsection{Row Distributed Ensemble}

In the row distributed ensemble framework, the target set $\mathcal{X}$ is partitioned into $D$ different subsets, 
which are assumed to be at different locations. The instances from subset $d$ are denoted by $\mathcal{X}_{d}$, so that 
$\mathcal{X}=\cup_{d=1}^{D}\mathcal{X}_{d}$. 
It is assumed that class and cluster labels are available -- 
\textit{i.e.}, they have already been generated by some classification and clustering algorithms. 
%(or by some distributed classification and clustering algorithms). 
The objective is to refine the class probability distributions (obtained from the classifiers) of the instances from $\mathcal{X}$ without sharing 
the class/cluster labels across the data sites. 
%The same privacy constraint w.r.t. the class/cluster labels holds if a central server is used. 

% \begin{center}
% \begin{figure}[ht]
% \centering
% \input{figrd.tex}
% \caption{Row Distributed Ensemble}
% \label{fig:rpart}
% \end{figure}
% \end{center}

% \begin{figure}[h]
%  \centering
%  \includegraphics[bb=0 0 210 204]{../Dr. Eduardo and Ayan/PASSAT11_08_28_2011/rdb.png}
%  % rdb.png: 280x272 pixel, 96dpi, 7.41x7.20 cm, bb=0 0 210 204
% \end{figure}
% 
% \begin{figure}[h]
%  \centering
%  \includegraphics[bb=0 0 205 204,scale=0.4]{../Dr. Eduardo and Ayan/PASSAT11_08_28_2011/cdb.png}
%  % cdb.png: 274x272 pixel, 96dpi, 7.25x7.20 cm, bb=0 0 205 204
% \end{figure}
% 
% \begin{figure}[h]
%  \centering
%  \includegraphics[bb=0 0 240 316,scale=0.5]{../Dr. Eduardo and Ayan/PASSAT11_08_28_2011/arbd.png}
%  % arbd.png: 320x421 pixel, 96dpi, 8.47x11.14 cm, bb=0 0 240 316
% \end{figure}

A careful look at the E-step -- Equations (\ref{updatephi}) and (\ref{updategamman}) -- reveals that the update of the variational parameters corresponding 
to each instance in a given iteration is independent of those of other instances given the model parameters from the previous iteration. 
This suggests that we can maintain a client-server based framework, where the server only updates the model parameters (in the M-step) and the 
clients (corresponding to individual data sites) update the variational parameters of the instances in the E-step. 
For instance, consider a situation (shown in Fig. \ref{fig:rpart}) where a target dataset $\mathcal{X}$ is partitioned into 
two subsets, $\mathcal{X}_{1}$ and $\mathcal{X}_{2}$, and that these subsets are located in two different data sites. The data site $1$ has access 
to $\mathcal{X}_{1}$ and accordingly, to the respective class and cluster labels of their instances. Similarly, the data site $2$ has access to the 
instances of $\mathcal{X}_{2}$ and their class/cluster labels. 
% Note that, in this case, a common set of classification and clustering algorithms has been 
% applied to both data sites. 

%This scenario is illustrated in Fig. \ref{fig:rpart}, where $G$ represents the set of classifiers and clusterers employed for both data sites. Note that, in this case, a common set of classification and clustering algorithms has been applied to both data sites. 

Now, data site $1$ can update the variational parameters
$\{\vect{\zeta}_{n}\}\text{ }\forall \mathbf{x}_{n}\in \mathcal{X}_{1}$. Similarly, data site $2$ can update the 
variational parameters $\{\vect{\zeta}_{n}\}\text{ }\forall \mathbf{x}_{n}\in \mathcal{X}_{2}$. Once the variational parameters are updated
in the E-step, the server gathers information from the two sites and updates the model parameters. 
Here, the primary requirement is that the class and cluster labels of instances from different data sites should not be available to the server. 
% A closer
% inspection of the M-step update (Eq. (\ref{updatebeta})) reveals that it contains a summation over the instances. 
% Therefore, individual data sites can send only some collective information to the server without transgressing privacy. 
Now, Eq. (\ref{updatebeta}) can be broken as follows:
\begin{equation}
\label{decomposition}
{\beta_{mij}}^{*}\propto\displaystyle\sum_{x_{n}\in\mathcal{X}_{1}}\phi_{nmi}w_{2nmj} + 
\displaystyle\sum_{x_{n}\in\mathcal{X}_{2}}\phi_{nmi}w_{2nmj}
\end{equation}
The first and second terms can be calculated in data sites $1$ and $2$, separately, and then sent to the server, where the two terms can be added 
and $\beta_{mij}$ can get updated $\forall m,i,j$. The variational parameters $\{\phi_{nmj}\}$ are not available to the sever and thus 
only some aggregated information about the values of $\{w_{2nm}\}$ for some $\mathbf{x}_{n}\in\mathcal{X}$ is sent to the server.
We also observe that more the number of instances in a given data site, more difficult it becomes to retrieve the cluster labels (\textit{i.e.} $\{w_{2nm}\}$)
from individual clients. Also, in practice, the server does not get to know how many instances are present per data site which only makes the recovery
of cluster labels even more difficult. Also note that the approach adopted only splits a central computation in multiple tasks based on how the data
is distributed. Therefore, the performance of the proposed model with all data in a single place should always be the same as the performance 
with distributed data assuming there is no information loss in data transmission from one node to another. 

In summary, the server, after updating $\vect{\zeta}_{0}$ in the M-step, sends them out to the individual clients.
The clients, after updating the variational parameters in the E-step, send some partial summation results in the form shown in
Eq. (\ref{decomposition}) to the server. The server node is helpful for the conceptual understanding of the parameter update and sharing procedures. 
In practice, however, there is no real need for a server. Any of the client nodes can 
itself take the place of server, provided that the computations are carried out in separate time windows and 
in proper order. 

\subsection{Column and Arbitrarily Distributed Ensemble}
The column and arbitrarily distributed ensembles are illustrated in Figs. \ref{fig:cpart} and \ref{fig:arbpart} respectively. 
Analogous distributed inference and estimation frameworks can be derived in these two cases without sharing the 
cluster/class labels among different data sites. However, detailed discussion is avoided due to space constraints.

% \scriptsize
% \begin{table*}[hbp]
% \centering
% \begin{tabular}{|l||c|c|c|c|c|}
% \hline
% \hline
% 
%    Dataset & Best Component    & Classifier Ensemble   & C\textsuperscript{3}E    & BGCM   & BC\textsuperscript{3}E\\ \hline
%    
%    Halfmoon (2\%) &$93.02\pm0.81$   &$92.53\pm1.83$   &$\textbf{99.64}\pm \textbf{0.08}$   &$99.58\pm 0.08$   &$99.37\pm 1.57$\\ \hline
% 
%    German Numer (10\%) &$68.20\pm 1.06$   &$67.70\pm 1.34$   &$\textbf{70.77}\pm \textbf{0.26}$   &$67.83\pm 1.74$   &$70.30\pm 1.17$\\ \hline
% 
%    Heart (7\%)   & $72.91\pm4.60$   &$71.26\pm 6.18$   &$83.54\pm3.79$    & $\textbf{84.54}\pm \textbf{2.63}$  & $82.96\pm 6.35$\\ \hline                                        
% 
%    Indian Pima (5\%) &$70.65\pm2.77$   &$69.95\pm 2.48$   &$74.75\pm 2.66$   &$\textbf{75.86}\pm \textbf{1.25}$   &$75.29\pm2.23$\\ \hline                    
%                     
%    Wine (10\%) &$84.32\pm5.62$   &$83.84\pm 6.29$   &$\textbf{88.79}\pm \textbf{3.52}$   &$85.43\pm 5.23$   &$88.60\pm3.82$\\ \hline           
%     
% \hline
% \hline
% \end{tabular}
% \caption{Experimental Evaluation of BC\textsuperscript{3}E}
% \label{tableresultsbc3e}
% \end{table*}
% \normalsize

\section{Experimental Evaluation}
\label{sec:eval}

We have already shown, theoretically, that the classification results obtained by the privacy-aware \textbf{BC\textsuperscript{3}E} are precisely the same as those we would have gotten if all the information originally distributed across different data sites were available at a single data site. 
Therefore, we assess the learning capabilities of \textbf{BC\textsuperscript{3}E} using 
five benchmark datasets (\textit{Heart}, \textit{German Numer}, \textit{Halfmoon}, \textit{Wine}, and \textit{Pima Indians Diabetes}) --- all stored in a single location. 
Semi-supervised approaches are most useful when labeled data is limited, while these benchmarks were created for evaluating supervised methods. Therefore, we use only small portions (from 2\% to 10\%) of the training data to build classifier ensembles. 
The remaining data is used as a target set --- with the labels removed. We adopt 3 classifiers (Decision Tree, Generalized Logistic Regression, and Linear Discriminant). For clustering, we use hierarchical single-link and $k$-means algorithms. 
% Considering \textbf{BC\textsuperscript{3}E}, we use random initialization for both $\alpha$ and $\beta$ and start with the M-step first (instead of the E-step), but ensure that the initial class assignments are highly skewed (e.g., 
% the scaling factor of the Dirichlet distribution for $\alpha$ is set to 0.9, which guarantees some sparsity). 
The achieved results are presented in Table I,
%\ref{tableresultsbc3e}
where \textit{Best Component} indicates the accuracy of the best 
classifier of the ensemble. We also compare \textbf{BC\textsuperscript{3}E} with two related algorithms (\textbf{C\textsuperscript{3}E} \cite{achg11} and \textbf{BGCM} \cite{galf09}) that do not deal with privacy issues. One can observe that, besides 
having the privacy-preserving property, \textbf{BC\textsuperscript{3}E} presents competitive accuracies with respect to their counterparts. Indeed, the Friedman test, followed by the Nemenyi post-hoc test for pairwise 
comparisons between algorithms, shows that there is no significant statistical difference ($\alpha=10\%$) among the accuracies of \textbf{BC\textsuperscript{3}E}, \textbf{C\textsuperscript{3}E}, and \textbf{BGCM}.

%The empirical performance of BC\textsuperscript{3}E is observed in comparison with existing approaches 
%on four different datasets selected from the repositories of \href{http://www.csie.ntu.edu.tw/~cjlin/libsvmtools/datasets/}{LIBSVM} 
%and \href{http://archive.ics.uci.edu/ml/}{UCI}. The proposed model, however, has some specialized application as mentioned before.
%If the classifier accuracies are too high, there is no real need to depend on clustering information. On the other hand, if the
%classification results are too poor, BC\textsuperscript{3}E cannot improve upon them as the evidences themselves are 
%corrupted. Hence, for proper working of the model, both the classification and clustering results have to be 
%moderately strong. This, in fact, is the ideal scenario for transfer learning where the classifiers trained on some older data
%are employed to classify some new target data where concept drift might have taken place. A similar strategy is adopted to
%simulate this situation with the existing datasets. From each dataset, only the minimal percentage of data is selected for training a classifier ensemble
%consisting of decision tree, generalized logistic regression and linear discriminant analysis. Rest of the data is treated as target data which is
%clustered using either hierarchical single-link agglomerative clustering or $k$-means clustering. 
%The results are presented in table \ref{tableresultsbc3e}. The column `max' indicates the maximum accuracy from a classifier and `avg' represents the
%accuracy from the classifier ensemble. 

\section{Extension and Future Work}
\label{conclusions}
The results achieved so far motivate us to employ soft classification and clustering. 
%which can leverage the accuracy of \textbf{BC\textsuperscript{3}E}. 
Applications of \textbf{BC\textsuperscript{3}E} to real-world transfer learning problems are also in order. 
% From the learning viewpoint, the development 
% of a Gibbs sampling based distributed algorithm is in progress. Although these are promising research venues, we believe that a more impactful contribution 
% would be studying a dynamic version of \textbf{BC\textsuperscript{3}E} by assuming that: (i) the data resides in distributed servers; and 
% (ii) the data characteristics across different sites change over time. For instance, if some bank branches
% %distributed over a given demographic region, 
% decide to track the temporal changes in the behavior of the customers, a dynamic version of \textbf{BC\textsuperscript{3}E} could be used 
% without any leakage of private information about the customers. Similarly, different e-commerce sites could 
% %share information about customers to
% track changes in customer behavior over time to better understand their needs without revealing information about individuals.
% %in their purchase behavior for better understanding of the users' needs. 
% %This model would be idle for this kind of application where individual-level information is not shared.

\section*{Acknowledgments}
This work was supported by NSF (IIS-0713142 and IIS-1016614) and by the Brazilian Agencies FAPESP and CNPq.

\bibliographystyle{plain}
\begin{small}
 \bibliography{bibfile_PASSAT}
\end{small}

\end{document}